\documentclass{ieeeaccess}
\usepackage{cite}
\usepackage{amsmath}
\usepackage{amssymb}
\usepackage{dsfont}
\usepackage{textcomp}
\def\BibTeX{{\rm B\kern-.05em{\sc i\kern-.025em b}\kern-.08em
    T\kern-.1667em\lower.7ex\hbox{E}\kern-.125emX}}

\usepackage{graphicx}
\usepackage{array}
\usepackage{multirow}
\usepackage{adjustbox}
\usepackage{dsfont}
\usepackage{color,soul} 
\usepackage{scalerel}
\usepackage{makecell}

\usepackage{algorithm}
\usepackage[noend]{algpseudocode}

\newcommand{\argmin}{\arg\!\min} 
\newcommand\Tstrut{\rule{0pt}{2.6ex}}         
\newcommand\Bstrut{\rule[-0.9ex]{0pt}{0pt}}   

\usepackage[utf8]{inputenc}
\inputencoding{utf8}
\usepackage[colorlinks=false, hidelinks]{hyperref}
\usepackage{xcolor}
\definecolor{blue2}{HTML}{00AEEF}
\definecolor{blue1}{HTML}{0B6FAE}
\usepackage{lineno}

\begin{document}
	
\history{}
\doi{10.1109/ACCESS.2021.3124760}

\title{Strengthening Probabilistic Graphical Models: The Purge-and-merge Algorithm}
\author{\uppercase{Simon Streicher}\href{https://orcid.org/0000-0001-9281-0411}{\includegraphics[scale=0.047]{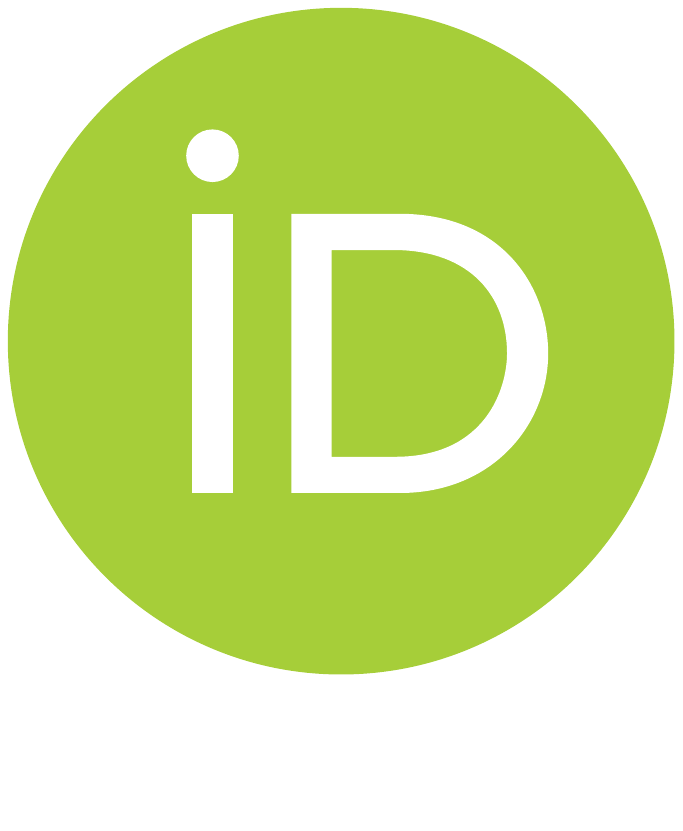}} and \uppercase{Johan du Preez}}

\address[]{Department of Electrical and Electronic Engineering, Stellenbosch University,	Stellenbosch 7600, South Africa (e-mail: dupreez@sun.ac.za)}

\markboth
{Simon Streicher and Johan du Preez. \,Strengthening Probabilistic Graphical Models: The Purge-and-merge Algorithm}
{Simon Streicher and Johan du Preez. \,Strengthening Probabilistic Graphical Models: The Purge-and-merge Algorithm}

\corresp{Corresponding author: Simon Streicher (e-mail: sfstreicher@gmail.com).}

\begin{abstract}
	Probabilistic graphical models (PGMs) are powerful tools for solving systems of complex relationships over a variety of probability distributions. However, while tree-structured PGMs always result in efficient and exact solutions, inference on graph (or loopy) structured PGMs is not guaranteed to discover the optimal solutions~\cite[p391]{koller}. It is in principle possible to convert loopy PGMs to an equivalent tree structure, but this is usually impractical for interesting problems due to exponential blow-up~\cite[p336]{koller}. To address this, we developed the ``purge-and-merge'' algorithm. This algorithm iteratively nudges a malleable graph structure towards a tree structure by selectively \textit{merging} factors. The merging process is designed to avoid exponential blow-up by way of sparse structures from which redundancy is \textit{purged} as the algorithm progresses. We set up tasks to test the algorithm on constraint-satisfaction puzzles such as Sudoku, Fill-a-pix, and Kakuro, and it outperformed other PGM-based	approaches reported in the literature~\cite{BaukeH, GoldbergerJ, KhanS}. While the tasks we set focussed on the binary logic of CSP, we believe the purge-and-merge algorithm could be extended to general PGM inference.
\end{abstract}


\titlepgskip=-15pt

\maketitle

\section{Introduction}
We have successfully created flexible probabilistic graphical model (PGM) structures to solve constraint-satisfaction problems (CSPs) that cannot be solved with existing PGM inference techniques. This entailed the creation of an exact CSP solver that preserves all solutions. 

We did not set out to explore modern constraint-satisfaction problem solving in general, but rather to incorporate constraint-satisfaction capabilities into PGMs. Central to this work is a PGM technique called purge-and-merge. It is the combination of three established probabilistic techniques: building cluster graphs~\cite{streicher}, applying loopy belief propagation~\cite{dechter2010on}, and merging factors into a joint space. Together, these techniques enable purge-and-merge to allow the growth of factors via factor merging while also removing redundancies in the CSP problem space via loopy-belief propagation. We can thus solve a range of CSPs that would be too intricate for either loopy-belief propagation or factor merging. Our experimental study shows that purge-and-merge reliably solves problems too difficult for other belief-propagation approaches~\cite{BaukeH, GoldbergerJ, KhanS, streicher}.

Purge-and-merge provides higher-order reasoning for PGMs and constraint satisfaction. This technique would therefore be of benefit to any area that incorporates both these domains, such as:
\begin{itemize}
	\item classification and re-classification problems -- e.g. image de-noising~\cite{koller} and scene classification~\cite{cvexample1};
	
	\item image segmentation -- e.g. breaking an image up into superpixels using pixel similarity and boundary constraints; and
	
	\item hybrid reasoning, -- e.g. solving Sudokus visually by combining a handwriting input classifier with constraint satisfaction.
\end{itemize}

PGMs are tools that express intricate problems with multiple dependencies as graphs. PGM inference techniques such as message passing can then be used to solve these graphs. PGMs are integral to a wide range of probabilistic problems~\cite{sucar2015probabilistic} such as medical diagnosis and decision making~\cite{medicalexample}, object recognition in computer vision~\cite{cvexample1}, as well as speech recognition and natural-language processing~\cite{nlpexample}.

Constraint satisfaction in turn is classically viewed as a graph search problem that falls under the umbrella of NP-complete problems. It originated in the artificial intelligence (AI) literature of the 1970s, with early examples in Mackworth~\cite{Mackworth} and Laurière~\cite{Lauriere}. Broadly, a CSP consists of a set of variables $\mathcal{X} = \{X_1, X_2, \ldots, X_N\}$, where each variable must be assigned a value such that a given set of constraints (clauses) $\mathcal{C} = \{C_1, C_2, \ldots, C_M\}$ are satisfied. Typical applications of CSPs include resource management and time scheduling~\cite{cspplanning}, parity checking in error-correcting codes~\cite{paritycheck}, and puzzle games such as Sudoku, Killer-Sudoku, Calcudoku, Kakuro, and Fill-a-pix~\cite{csppuzzle}.

Many advances have been made in solving highly constrained PGMs, i.e. PGMs with a large number of prohibited outcomes specified in their factors. This includes PGMs constructed from CSP clauses. A popular approach is to transform such a PGM to another domain and then to solve it with tools specific to that domain. This includes converting PGMs to Boolean satisfiability problems (such as conjugate normal form (CNF)~\cite{cnfbool}), sentential decision diagrams (SDD)~\cite{sddchoi}, and arithmetic circuits (ACs)~\cite{acewebsite}. For example, the ACE system~\cite{acewebsite} compiles a factor graph or Bayesian network into a separate AC. This AC is then used to answer queries about the underlying variables. The drawback lies in the fact that an ACE query will yield a marginal for each queried variable, but it does not yield the joint distribution over all queried variables. ACE thus acts as a heuristic for selecting a single probable CSP outcome. 

In contrast, our CSP solver can return a joint solution to a CSP problem (i.e. find its joint distribution) using PGM techniques.

Of course, there are trivial ways to reformulate CSPs probabilistically and express them as PGMs~\cite{MoonT, GoldbergerJ, KhanS, BaukeH, LakshmiA, streicher}. Although most of the above citations are aimed at specific CSPs, they share the same basic approach. This amounts to (a) formulating the CSP clauses into PGM factors, (b) configuring the factors into a PGM graph structure, (c) applying belief propagation on this graph, and (d) using the most probable outcome as the solution to the CSP. Dechter~\cite{dechter2010on} provides a bridge between CSPs and PGMs by proving that zero-belief conclusions made by loopy-belief propagation reduce to an algorithm for generalised arc consistency in CSPs.

There are limitations, however. Goldberger~\cite{GoldbergerJ} highlights the difference between belief propagation (BP) with max-product and sum-product. They report that although max-product BP ensures the solution is preserved at all times, it is often hidden within a large spectrum of possibilities. This calls for additional search techniques. Meanwhile, sum-product BP acts as a heuristic to highlight a valid solution, but can often highlight an incorrect one. Khan~\cite{KhanS} tries to improve on the success rate of sum-product BP by combining it with Sinkhorn balancing. Although they report an improvement, the system could still not reliably solve high-difficulty Sudoku puzzles. Streicher~\cite{streicher} use a sparse representation for factors and promote the use of a cluster graph over the ubiquitous factor graph. However, although the cluster graphs improved the accuracy and execution time of the system, their approach is not reliable as a Sudoku solver or CSP solver in general.

The above approaches are all limited in one way or another. They are either ineffective in purging redundant search space -- due to their loopy PGM structure -- or they rely on an unreliable heuristic to select a probable solution. In this work, we propose techniques to sidestep these limitations and iteratively nudge the graph towards a tree-structured PGM while preserving the CSP solution.

Our proposed technique employs the purge-and-merge algorithm. Purge-and-merge starts by constructing a CSP into a cluster graph PGM with sparse factors~\cite{streicher}. It then \textit{purges} redundancies from these factors by applying max-product belief propagation~\cite{dechter2010on} and thereby propagating zero-belief conclusions. Next, it \textit{merges} factors together to create cluster graphs that are closer to a tree structure. Finally, it constructs a new cluster graph from the factors. This process is repeated until a tree-structure cluster graph is produced. At this point, the exact solution to the CSP is found.

Purge-and-merge manages to reliably solve CSPs that are too difficult for the aforementioned approaches. We reason that a successful CSP approach such as purge-and-merge opens many new avenues for exploration in the PGM field. This may include hybrid models where rigid and soft constraints can be mixed. It may also be used in domains not previously suited for probabilistic approaches.

Our study is outlined as follows:
\begin{itemize}
	\item
	In Section~\ref{sec-CSP} we introduce CSP factors and show how they can be structured into a PGM. We provide the design and techniques to build and solve a basic constraint-satisfaction PGM.
	
	\item
	In Section~\ref{sec:limitationsofpgms} we investigate the limitations of PGMs as well as the trade-offs between the loopy-structured PGMs of small-factor scopes and the tree-structured PGMs of large-factor scopes.
	
	\item
	In Section~\ref{sec:purgeandmerge-main} we provide a factor clustering and merging routine along with the purging methods found in the purge-and-merge technique.
	
	\item
	In Section~\ref{sec:experiments} we evaluate purge-and-merge on a number of example CSPs such as Fill-a-pix and similar puzzles, and compare it to the ACE system~\cite{acewebsite}.
\end{itemize}

We found that with the purge-and-merge technique, PGMs can solve highly complex CSPs. We therefore conclude that our approach is successful as a CSP solver, and suggest further investigation into integrating constraint-satisfaction PGMs as sub-components of more general PGMs.

\section{Constraint satisfaction using PGMs}\label{sec-CSP}

In this section we show how CSPs are related to PGMs. We express CSPs as factors, which can be linked in a PGM structure. We use graph colouring as an example, and expand the idea to the broader class of CSPs.

Most constraint-satisfaction problems are easily defined and verified, but they can be difficult to invert and solve. PGMs, by contrast, are probabilistic reasoning tools used to resolve large-scale problems in a computationally feasible manner. They are often useful for problems that are difficult to approach algorithmically -- CSPs being one such example.

\subsection{A general description of CSPs}
Constraint-satisfaction problems are NP-complete. They are of significant importance in operational research and they are key to a variety of combinatorial, scheduling, and optimisation problems.

In general, constraint satisfaction deals with a set of variables $\mathcal{X} = \{X_1, X_2, \ldots, X_N\}$ and a set of constraints $\mathcal{C} = \{C_1, C_2, \ldots, C_M\}$. Each variable needs to be assigned a value from the variable's finite domain $\text{dom}(X_n)$, such that all constraints are satisfied. For example, if $X_n$ represents a die roll, then a suitable domain would be $\{1,2,3,4,5,6\}$. Furthermore, if we define a CSP where two die rolls, $X_1$ and $X_2$, are constrained to sum to a value of $10$, then the CSP solution $(X_1, X_2)$ consists of the possible value assignments $(4,6)$, $(5,5)$, and $(6,4)$.

The CSP constraints can be visualised through a factor graph. This is a bipartite graph where the CSP variables are represented by variable nodes (circles) and the CSP clauses by factor nodes (rectangles). The edges of the graph are drawn between factor nodes and variable nodes, such that each factor connects to all the variables in its scope. The scope of a factor is the set of all random variables related to that factor. To illustrate, we present the map-colouring example in Figure~\ref{fig-mapcolouring}:
\begin{itemize}
	
	\item
	Figure~\ref{fig-mapcolouring}(a) shows a map with bordering regions. These regions are to be coloured using only four colours such that no two bordering regions may have the same colour.
	
	\item
	In Figure~\ref{fig-mapcolouring}(b) we represent this map as a graph-colouring problem, where the regions are represented by nodes and the borders by edges.
	
	\item
	Figure~\ref{fig-mapcolouring}(c) shows a factor graph where the factors represent the CSP clauses. Note that each of these factors has a scope of two variables.
	
	\item
	In Figure~\ref{fig-mapcolouring}(d) we show that the problem can also be expressed equivalently by combining factors differently. Here we have multiple constraints captured by a single clause. As a result, we have fewer factors, but larger factor scopes. (This example uses the maximal cliques in (b) as factors.)
	
\end{itemize}

\subsection{Factor representation}

\begin{figure}[h!]
	\centering
	\includegraphics[width=0.98\columnwidth]{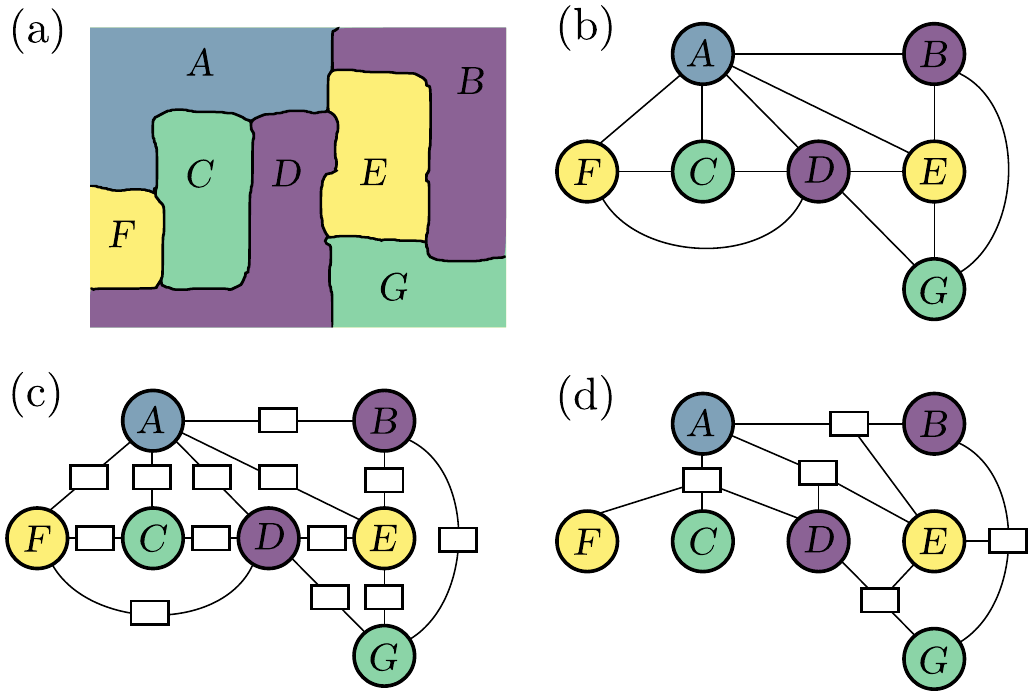}
	\caption{ (a) A map-colouring example, with (b) its graph-colouring representation, and (c) and (d) two different factor-graph representations using rectangles to represent CSP clauses.
	}\label{fig-mapcolouring}
\end{figure}

\begin{figure}[h]
	\centering
	\hspace{-1em}\includegraphics[scale=0.785]{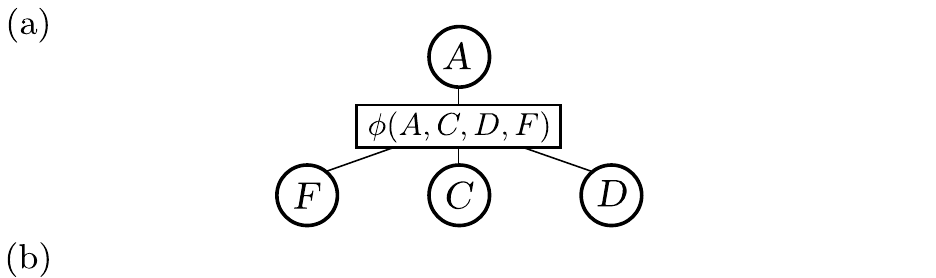}
	\\$\ $\\
	\vspace{-0.82em}
	\begin{tabular}{r c c c c |c }
		~& $A$ & $C$ & $D$ & $F$ & $\phi(A,C,D,F)$  \\
		\cline{2-6}
		& 1\Tstrut & 2 & 3 & 4 & 1 \\
		& 1 & 2 & 4 & 3 & 1 \\
		& 1 & 3 & 2 & 4 & 1 \\
		& 1 & 3 & 4 & 2 & 1 \\
		& $\vdots$ & $\vdots$ & $\vdots$ & $\vdots$ & $\vdots$ \\
		& 4 & 3 & 2 & 1 & 1 \\
		\cline{2-6}
		& \multicolumn{4}{c|}{elsewhere\Tstrut} & 0 
	\end{tabular}
	\vspace{0.1em}
	\caption{(a) The map-colouring clause $\{A,B,D,F\}$ from Figure~\ref{fig-mapcolouring}, with factor $\phi(A, C, D, F)$, variables $A$,  $C$, $D$, $F$, and variable domains $\text{dom}(A)$ = $\text{dom}(B)$ = $\text{dom}(C)$ = $\text{dom}(D)$ = $\text{dom}(F)$ = $\{1,2,3,4\}$. (b) A sparse table explicitly listing the non-zero entries in $\phi(A, C, D, F)$, and assigning all other entries to be zero.}
	\label{fig-probtable}
\end{figure}

A factor graph representation will only show the clauses, the variables, and the relationship between the clauses and variables. The details of these relationships, however, are suppressed. To fully represent the underlying CSP, each factor must also express the relationships implied by the associated constraint. We do so by assigning a potential function to each clause in order to encode all valid local assignments concerning that clause. These assignments are captured by sparse probability tables. The tables list each local possibility as a potential solution, and assign a value to that possibility. For CSPs specifically, we work with binary probabilities, ascribing ``1'' to any (valid) possibility and ``0'' to any impossibility enforced by the constraint. As an example, see Figure~\ref{fig-probtable} for the sparse table representing the factor scope $\{A,C,D,F\}$ from Figure~\ref{fig-mapcolouring}(d). (The use of sparse tables as PGM factors is also referred to as flattening~\cite{dechter2010on}.)

It is worth noting that the factor graphs presented here are then not only a visually appealing representation for CSPs, but are in fact PGMs. As such, PGM inference techniques such as loopy belief propagation~\cite{MoonT} and loopy belief update~\cite{koller, lauritzen1988local} can be directly applied to these factor graphs. 

In order to perform belief propagation using sparse tables, it is important to introduce some basic factor operations; most importantly, see the literature on factor multiplication, division, marginalisation, reduction, damping, and normalisation~[\citenum{koller}; Defs. 4.2, 10.7, 13.12, and 4.5; Eq. 11.14; and Ch. 4].

\subsection{PGM construction}\label{sec:ltrip}
In essence, a PGM is a compact representation of a probabilistic space as the product of smaller, conditionally independent distributions. When we apply a PGM to a specific problem, we need to (a) obtain factors to represent these distributions, (b) construct a graph from them, and (c) use inference on this graph. In this section we focus on graph construction.

A cluster graph is an undirected graph where:
\begin{itemize}
	\item each node $i$ is associated with cluster $C_i \subseteq \mathcal{X}$, 
	\item each edge $(i, j)$ is associated with a separation set (sepset) $S_{i,j} \subseteq C_i \cap C_j$, and 
	\item the graph configuration satisfies the running intersection property (RIP)~\cite{koller}
	.
\end{itemize}
RIP requires that for all pairs of clusters containing a common variable, $X \in C_i$ and $X \in C_j$, there must be a unique path of edges, $(\hat{\imath}, \hat{\jmath})$, between $C_i$ and $C_j$ such that $X \in S_{\hat{\imath}, \hat{\jmath}}\, \forall\, (\hat{\imath}, \hat{\jmath})$.

\begin{figure}[h]
	\centering
	\includegraphics[width=\linewidth]{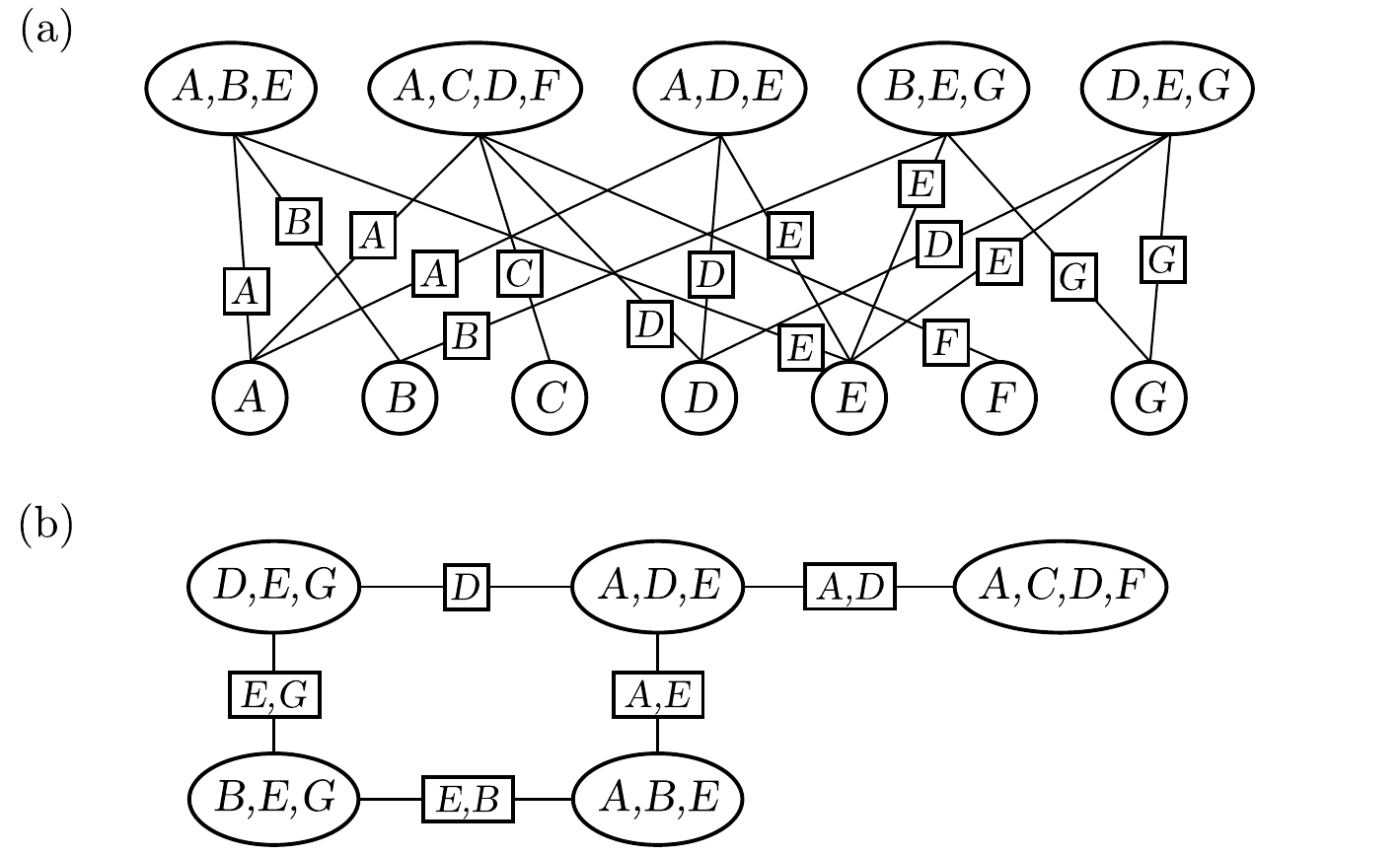}
	\caption{Two different cluster-graph configurations for the map-colouring example in Figure~\ref{fig-mapcolouring}. (a) A Bethe-graph configuration that satisfies RIP by connecting all CSP clauses via single-variable sepsets to singe-variable clusters. (b) A graph configuration generated by LTRIP with fewer edges and multivariate sepsets}
	\label{fig:betheltrip}
\end{figure}

Figure~\ref{fig:betheltrip} provides two examples of a cluster-graph configuration for the CSP clauses in Figure~\ref{fig-mapcolouring}. In (a) we have a trivial connection called a Bethe graph. This is a cluster graph with univariate sepsets, an equivalent of the factor graph in Figure~\ref{fig-mapcolouring}(d). In (b) we have a cluster graph with multivariate sepsets. This graph is generated from the same factors as the Bethe graph, but using the LTRIP algorithm\cite{streicher}. The result is also referred to as a junction graph~\cite{dechter2010on}.

Streicher~\cite{streicher} shows that cluster graphs with multivariate sepsets have superior inference characteristics to factor graphs, both in terms of both speed and accuracy. The same is argued by Koller~\cite[Sec. 11.3.5.3]{koller}, where it is shown that cluster graphs are a more general case of factor graphs without the limitation of passing messages only through univariate marginal distributions. With factor graphs, correlations between variables are lost during belief propagation, which can have a negative impact on the accuracy of the posterior distributions and on the number of messages required for convergence.

The LTRIP algorithm is designed to configure factors into a valid cluster graph by following the RIP constraints. For each variable $X \in \mathcal{X}$, LTRIP builds a subgraph out of all clusters $C_i$, where $X \in C_i$. These subgraphs are then superimposed in order to construct the sepsets of the final graph. In summary, the algorithm states that for each variable $X \in \mathcal{X}$, do the following:
\begin{itemize}
	\item find all clusters $C_i$ such that $X \in C_i$,
	\item construct a complete graph over clusters $C_i$,
	\item assign edge weights $w_{\hat{\imath}, \hat{\jmath}}$ to represent the similarity between neighbouring clusters,\footnote{The source implementation uses cluster intersections as edge weights: $w_{\hat{\imath}, \hat{\jmath}} = |C_{\hat{\imath}} \cup C_{\hat{\jmath}}|$~\cite{streicher}. Other suggestions include mutual information or the entropy over the shared variables.}
	\item connect the graph into a minimum spanning tree by using an algorithm such as the Prim-Jarn\'{i}k algorithm~\cite{prim}, and
	\item populate the tree's edges with intermediate sepset results $S^X_{\hat{\imath}, \hat{\jmath}} = \{X\}$.
\end{itemize}
After the sepset results are populated for each variable, the sepsets $S_{\hat{\imath}, \hat{\jmath}}$ of the final graph are taken as the union of the intermediate sepset results  $S_{\hat{\imath}, \hat{\jmath}} = \cup_{X \in \mathcal{X}} S^X_{\hat{\imath}, \hat{\jmath}}$.

\begin{figure}[h]
	\centering
	\includegraphics[width=\linewidth]{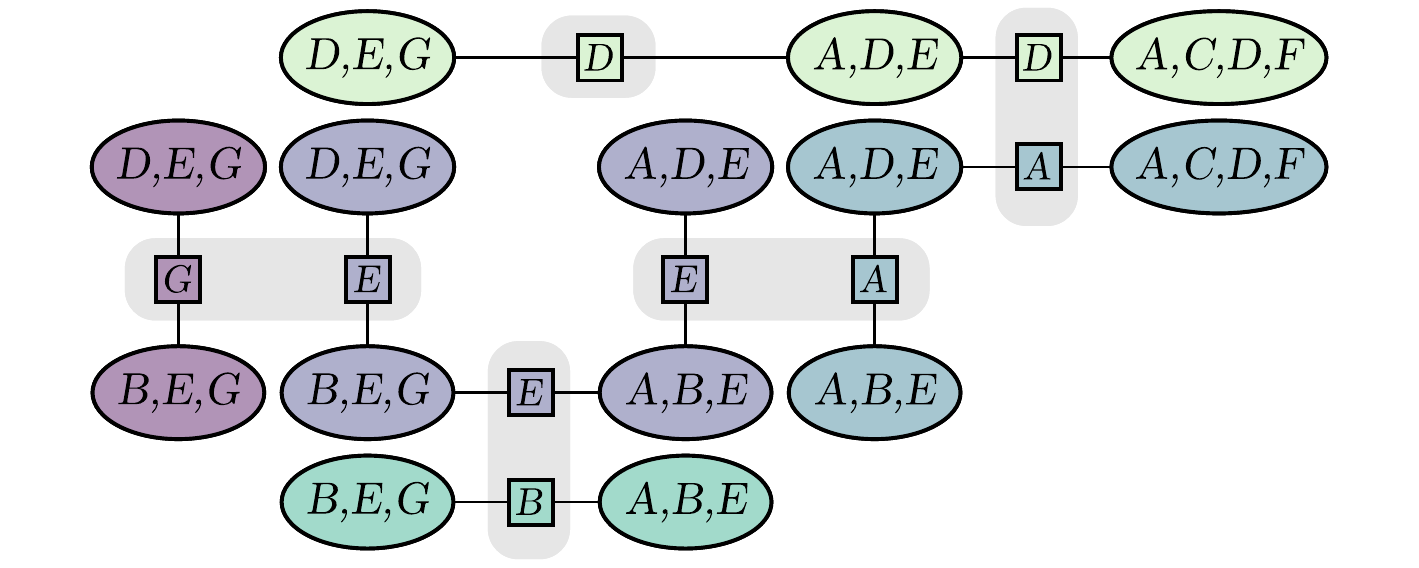}
	\caption{An example of applying the LTRIP algorithm in order to achieve the cluster-graph construction from Figure~\ref{fig:betheltrip}(b). For each variable $A, B, C, D, E, F$ and $G$, a minimum spanning tree is constructed from its associated clusters and is populated with univariate sepsets. The resulting cluster graph is then created by taking the superposition of these intermediate trees.}
	\label{fig:ltrip_construction}
\end{figure}

An example of the LTRIP algorithm for the graph in Figure~\ref{fig:betheltrip}(b) can be seen in Figure~\ref{fig:ltrip_construction}.

\subsection{PGM inference}\label{sec:basig_pgm}
Our PGM approach extends the prior work of Streicher~\cite{streicher}, which in turn is informed by the work of Dechter~\cite{dechter2010on}. The specific design choices for our PGM implementation are as follows:
\begin{enumerate}
	\item
	The factors consist of sparse tables similar to those of Figure~\ref{fig-probtable}.
	
	\item
	Graphs construction is done using the LTRIP procedure from Streicher~\cite{streicher}.
	
	\item
	We use inference via belief \emph{update} (BU) message passing, also known as the Lauritzen-Spiegelhalter algorithm~\cite{lauritzen1988local}.
	
	\item
	We use the Kullback-Leibler divergence as a comparative
	metric (and deviation/error metric) between distributions.
	
	\item
	Message-passing schedules are set up according to residual belief propagation~\cite{elidan2012residual}. Messages are prioritised according to the deviation between a new message and the preceding message at the
	same location within the graph.
	
	\item
	Convergence is reached when the largest message deviation falls below a chosen threshold.
	
	\item
	Throughout the system, we use max-normalisation and max-marginalisation, as opposed to their summation equivalents.
\end{enumerate}

Dechter~\cite{dechter2010on} proved that zero-belief conclusions made by loopy belief propagation are correct and equal to inducing arc consistency. This is true in the case of using both sum or max operations~\cite{GoldbergerJ, streicher, dechter2010on}. This means that the basic PGM approach does not guarantee a solution to the CSP, but it does guarantee that all possible solutions are preserved.

We found convergence to be faster with the max operations than with sum operations. Furthermore, the max operations maintain a unity potential for all non-zero table entries. This is in line with the constraint-satisfaction perspective, where outcomes are either possible or impossible. Alternatively, if one is interested in a more dynamic distribution, the sum operations provide varying potentials that can be used as likelihood estimations~\cite{GoldbergerJ}.

Lastly, note that alternative message-passing techniques such as warning propagation and survey propagation~\cite{braunstein2005survey} are available. These two approaches attempt to elevate the solution from the problem space, but cannot guarantee that the solution is retained. Our interest is in pursuing an approach where the full solution space is preserved.

\section{The limitations of PGMs}\label{sec:limitationsofpgms}

\begin{figure}[h]
	\centering
	\hspace{-1em}\includegraphics[scale=0.785]{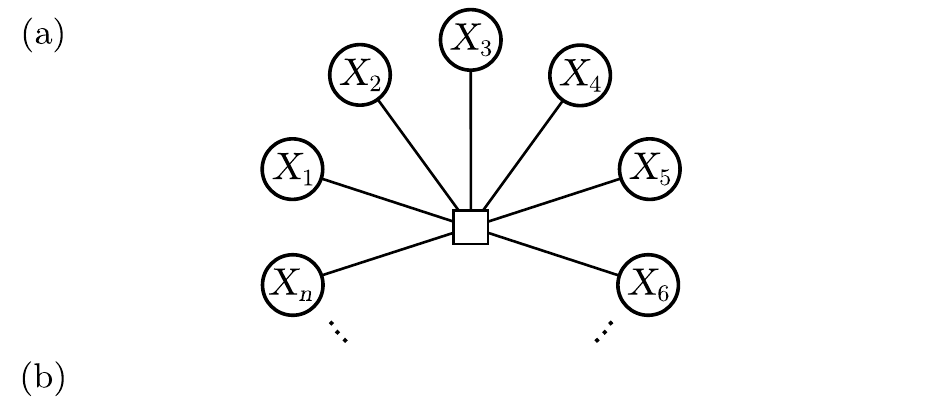}
	\\$\ $\\
	\vspace{-0.82em}
	\begin{tabular}{ c c c c c |c }
		$X_1$ & $X_2$ & $\cdots$ & $X_{n-1}$ & $X_n$ & $\phi(X_1,X_2,\cdots)$  \\
		\cline{1-6}
		1\Tstrut & 2 & $\cdots$ & n-1 &  n & 1 \\
		1 & 2 & $\cdots$ & n & n-1 & 1 \\
		$\vdots$ & $\vdots$ & $\ddots $ & $\vdots$  & $\vdots$ & $\vdots$ \Bstrut  \\
		n & n-1 & $\cdots$ & 2 & 1 & 1 \\
		\cline{1-6}
		\multicolumn{5}{c|}{elsewhere\Tstrut} & 0
	\end{tabular}
	\vspace{0.1em}
	\caption{ An example of an ill-suited problem with (a) its factor graph and (b) its sparse table containing $n!$ entries.}\label{fig-factorlimit}
\end{figure}

One of the main limitations of constructing CSP potential functions is the resources required to encode them. If the potential functions are encoded as probability tables, then at least all the non-zero potentials need to be listed. Such a list can grow exponentially with the number of factor variables. Therefore, not all CSPs are suitable to be expressed as sparse tables. A trivial example of an ill-suited problem would be a graph-colouring problem with $n$ fully connected nodes, as in Figure~\ref{fig-factorlimit}. The full space of the problem is $n^n$ with $n!$ entries in the probability table. 

Inference on loopy graphs is non-exact; it cannot guarantee a complete reduction to the solution space of a CSP~\cite{dechter2010on}. In exchange, however, loopy graphs provide a great advantage: the ability to handle problems that would have required infeasibly large probability tables if constructed into tree-structured PGMs.

\begin{figure}[!h]
	\centering
	\includegraphics[width=1\columnwidth]{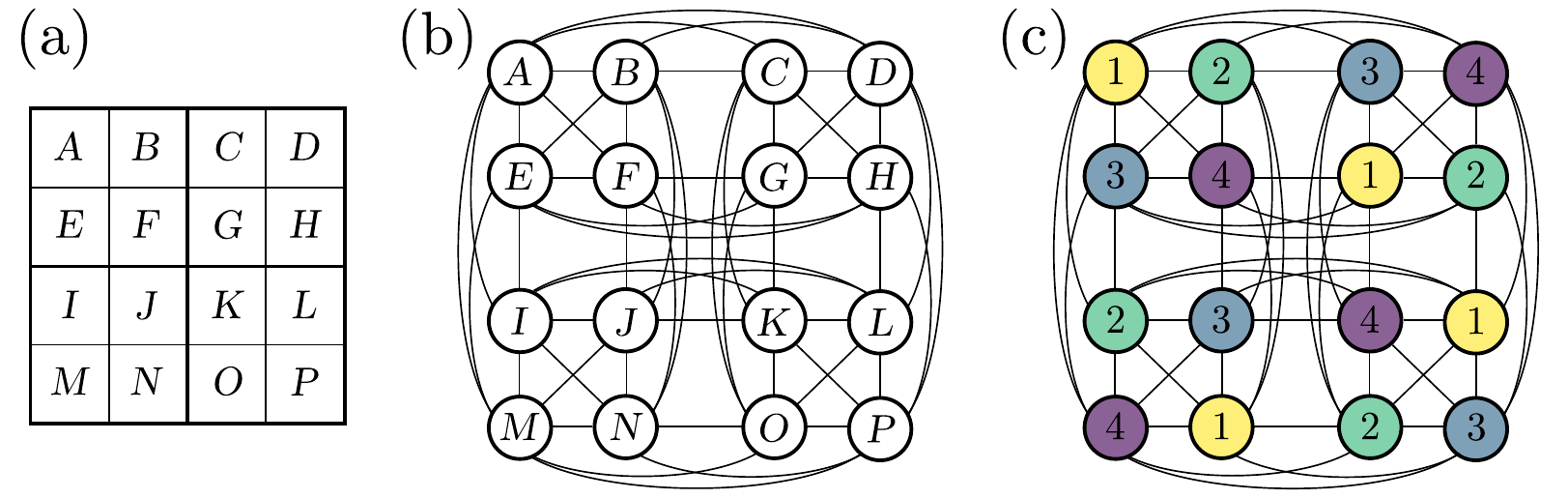}
	\caption{An example of a Sudoku puzzle as a graph-colouring problem: (a) is a $4\times 4$ version of a Sudoku puzzle, (b) connects the Sudoku variables in an undirected graph, and (c) shows one solution (of many) to this particular problem.
	}
	\label{fig:sudoku_colourgraph}
\end{figure}

Consider the Sudoku puzzle. A player is presented with a $9\times 9$ grid (with 91 variables) where each variable may be assigned a value of ``1'' to ``9'' and is constrained by the following 29 clauses: each row, each column, and each 3x3 non-overlapping subgrid may not contain any duplicates. Furthermore, a valid puzzle is partially filled with values such that only one solution exists. In Figure~\ref{fig:sudoku_colourgraph} we show the Sudoku-puzzle constraints as a graph-colouring problem.

Since a valid Sudoku should only have a single solution, the posterior distribution, $p(X_1, X_2, \ldots, X_{81})$, can be expressed by a sparse table that covers the full variable scope and contains only a single entry. Each marginal distribution, $p(X_1,\ldots,X_9)$, $p(X_{10},\ldots,X_{18})$, $\dots$, would then also hold only a single table entry. Yet, after the pre-filled values are observed and each factor is set up according to its local constraint (the no-duplicated rule), the prior distributions can result in tables that have as many as $9!$ entries. Therefore, a Sudoku solver would have to reduce these large initial probability tables into single-entry posteriors. Before we attempt such a solver, let us first consider two cases -- (1) a loopy structure with small-factor scopes, and (2) a tree structure with large-factor scopes:
\begin{enumerate}
	
	\item
	For model (1), we build a loopy-structured PGM directly from the prior distributions. Each factor, therefore, has the same variable scope as one of the clauses. For an inference attempt to be successful, factors should pass information around until all sparse tables are reduced to single entries. In practice, however, the sparse tables are often reduced by very little. This is because inference on a loopy structure does not guarantee convergence to the final solution space~\cite{dechter2010on}.
	
	\item
	For model (2), we use the most trivial tree structure: multiply all factors together to form a structure with a single node. The resulting factor will now contain one single table entry as the solution. This approach will often (or rather usually) fail in practice, since we cannot escape exponential blow-up. In the process of multiplying factors together, the intermediate probability tables first grow exponentially large before the system settles on this single-entry solution.
	
\end{enumerate}

Since we are confronted by the limitations of both small- and large-factor scopes, we propose a technique in the next section that mitigates these limitations.

\section{Purge-and-merge}\label{sec:purgeandmerge-main}
In this section, we consider the various methods for purging factors and merging factors, and combine these methods into a technique called purge-and-merge. It concludes with a detailed outline of the technique in Algorithm~\ref{alg:purgeandmerge}.

\subsection{Factor merging}\label{sec:factorclustering}
Our aim in merging factors is to build tree-structured PGMs and to be able to perform exact inference. This can result in exponentially larger probability tables, so it is necessary to approach this problem carefully.

One approach is to cluster the factors into subsets that will merge to reasonably sized tables. To pre-calculate the table size of a factor product is, unfortunately, as memory-inefficient as performing the actual product operation. While we therefore cannot use exact table size as a clustering metric, we have investigated three alternative metrics. We propose: (1) variable overlap, as in the number of overlapping variables between factors, (2) an upper-bound shared entropy metric, and (3) a gravity analogy that is built on entropy metrics. These methods are experimentally tested in Section{~\ref{sec:experiments}} (Figure{~\ref{fig-histogram}), with the gravity analogy showing the most potential.

\subsubsection{Variable overlap}
	For variable overlap, we define the attraction between $f_i$ and $f_j$ as
	\begin{equation*}
		a_{i, j} = \left| \mathcal{X}_i \cap \mathcal{X}_j \right|,
	\end{equation*}
	with $\mathcal{X}_i$ and $\mathcal{X}_j$ the scopes of $f_i$ and $f_j$ respectively. Note that there is a symmetrical relationship in the sense that the attraction of  $f_j$ towards $f_i$ can be defined as $a_{i \leftarrow j} = a_{i,j} = a_{j \leftarrow i}$.

\subsubsection{Upper-bound shared entropy}
	Upper-bound shared entropy is proposed as an alternative metric to variable overlap. The definition for the entropy of a set of variables $\mathcal{X}$ is
	\begin{equation*}
		H(\mathcal{X}) = \sum_{x \, \in \, \text{domain}(\mathcal{X})} -p(x) \log_2 p(x),
	\end{equation*}
	with a maximum upper bound achieved at the point where the distribution over $\mathcal{X}$ is uniform. This upper bound is calculated as
	\begin{equation*}
		\hat{\text{H}}\left(\mathcal{X}\right) = \log_2\left| \text{domain}(\mathcal{X}) \right|.
	\end{equation*}

	We use this definition to define the attraction between clusters $i$ and $j$ as the upper-bound entropy of the variables they share:
	\begin{equation*}
		a_{i, j} = \hat{\text{H}}(\mathcal{X}_i \cap \mathcal{X}_j),
	\end{equation*}
	with symmetrical behaviour $a_{i,j} = a_{i \leftarrow j} = a_{j \leftarrow i}$. Note that maximal entropy is used as a computationally convenient proxy for entropy, since calculating shared entropy directly is as expensive as applying factor product.

\subsubsection{Gravity method}
	For the gravity method, we use gravitational pull as an analogy for attraction:
	\begin{equation*}
		a_{i\leftarrow j} \propto m_i/ r_{i,j}^2.
	\end{equation*}

	The idea is to relate mass $m_i$ to how informed a factor is about its scope, and distance $r_{i,j}$ to concepts regarding shared entropy:

		
	\vspace{0.5em}$\hspace{-1em}$\textbf{Pseudo-mass:}\, 
	Mass equation $\text{m}(f) = m$  is based on how informed factor $f$ is about its scope $\mathcal{X}$. To parse this into a calculable metric we use the Kullback–Leibler divergence of the distribution of $f$ compared to a uniform distribution over $\mathcal{X}$.
	
	\vspace{0.5em}$\hspace{-1em}$\textbf{Pseudo-distance:}\, 
	As a distance metric, we want to register two factors as \textit{close together} if they have a large overlap and \textit{far apart} if they have little overlap. We also do not want this metric to be influenced by a factor's size or mass. 

	We arrived at a metric using the entropy of the joint distribution, normalised by the entropy of the variables shared between the factors. By using upper-bound entropy in our calculations, we arrive at distance
	\begin{equation*}
		r_{i,j} = \log_2\left(\frac{\hat{\text{H}}(\mathcal{X}_i \cup \mathcal{X}_j)}{\hat{\text{H}}(\mathcal{X}_i \cap \mathcal{X}_j)}\right).
	\end{equation*}
	
	\vspace{0.3em}$\hspace{-1em}$\textbf{Attraction:}\, 
	Finally, we define the attraction of $f_j$ towards $f_i$ as analogous to acceleration
	\begin{equation*}
		a_{i\leftarrow j} = \frac{m_i}{r^2_{i, j}}.
	\end{equation*}

	Using the above metrics, we formulate a procedure for clustering our factors according to the mergeability between factors, as shown in Algorithm~\ref{alg:clustering}. Although the algorithm is specialized for the gravity method, it can easily be adjusted for different attraction metrics.

	\begin{algorithm}[h!]
		\caption{\ Factor Clustering}
			{\bf Input:} factors $f_1, \ldots, f_n$ and threshold $\hat{H}_\tau$.\\
			{\bf Output:} clustered sets of factors $\mathcal{C}_1, \ldots, \mathcal{C}_m$, with property 
			\\${}$\hspace{3.49em} $\hat{\text{H}}(\text{vars in }\mathcal{C}_i) \leq  \hat{H}_\tau$ for all $\mathcal{C}_i$.
		\vspace{-0.5em}
		\\
		\rule{\columnwidth}{0.45pt}
		\begin{algorithmic}[1]
			
			\State {\color{blue1}// Initialize clusters and attractions}
			\For {each factor index $i$}
			\State $\mathcal{C}_i$ := $\{f_i\}$
			\State $\mathcal{X}_i$ := variables of $f_i$
			\State $m_i$ := m$(f_i)$
			\EndFor
			\For {each $i,j$ pair where $\left| \mathcal{X}_i \cup \mathcal{X}_j \right| > 0$}
			\State $a_{i\leftarrow j}$ := ${m_i}/{\text{r}(\mathcal{X}_i, \mathcal{X}_j)}$
			\State $a_{j\leftarrow i}$ := ${m_j}/{\text{r}(\mathcal{X}_j, \mathcal{X}_i)}$
			
			\EndFor
			\State {\color{blue1}// Dynamically merge clusters together}
			\While {any $a_{i \leftarrow j}$ are still available}
			\newcommand*{\ii}{\hat{\imath}} 
			\newcommand*{\jj}{\hat{\jmath}} 
			\State $\ii, \jj$ := $\argmin_{i, j}(a_{i \leftarrow j})$
			\If {$\hat{\text{H}}(\mathcal{X}_{\ii} \cup \mathcal{X}_{\jj}) \leq \hat{H}_\tau$}
			\State $\mathcal{C}_{\ii}$ := $\mathcal{C}_{\ii} \cup \mathcal{C}_{\jj}$
			\State $\mathcal{X}_{\ii}$ := $\mathcal{X}_{\ii} \cup \mathcal{X}_{\jj}$
			\State $m_{\ii}$ :=  $m_{\ii} + m_{\jj}$
			\For{each $k\hspace{-0.17em}\neq\hspace{-0.17em}{\jj}$ where $\left| \mathcal{X}_{\ii} \cup \mathcal{X}_k \right|>0$}
			\State $a_{{\ii}\leftarrow k}$ := ${m_{\ii}}/{\text{r}(\mathcal{X}_{\ii}, \mathcal{X}_k)}$
			\State $a_{k\leftarrow {\ii}}$ := ${m_k}/{\text{r}(\mathcal{X}_k, \mathcal{X}_{\ii})}$
			\EndFor
			
			\State remove
			$\mathcal{C}_{\jj}$,
			$\mathcal{X}_{\jj}$ and
			$m_{\jj}$
			\State remove $a_{{\jj} \leftarrow l}$ and
			$a_{l \leftarrow {\jj}}$, for any index $l$
			\Else \textbf{\ then}
			\State remove $a_{\ii \leftarrow \jj}$
			\EndIf
			\EndWhile
			\State
			\Return all remaining $\mathcal{C}_i, \mathcal{C}_j, \ldots$
		\end{algorithmic}\label{alg:clustering}
	\end{algorithm}
	
	Via this procedure, we can cluster factors $f_1, f_2, \ldots, f_n$  into clusters $\mathcal{C}_1, \ldots, \mathcal{C}_m$, where $m \leq n$. These clusters can then be incorporated into a PGM by calculating new PGM factors $f'_1, \ldots, f'_m$, by simply merging each cluster $f'_i = \prod_{f_j \in \mathcal{C}_i}f_j$.
	
	\subsection{Factor purging}\label{sec:factorpurging}
	In this section, we show some methods for purging the probability tables of a constraint-satisfaction PGM. We use the inference techniques from Section~\ref{sec:basig_pgm} along with some additional purging techniques:	
		
	\vspace{0.5em}$\hspace{-1em}$\textbf{Reducing variables:}\,  
	If for any factor $f_j$, a variable $X$ is uniquely determined to be $x_i$, i.e.\ there are no non-zero potentials with $X\neq x_i$ in that factor, then observe $X{=}x_i$ throughout all factors and remove $X$ from their scopes. This is a trivial case of ``node-consistency''~\cite{constraintnetworks}.
	
	\vspace{0.5em}$\hspace{-1em}$\textbf{Reducing domains:}\, 
	Likewise, if any domain entry $x_i \in \text{dom}(X)$ is not represented by factor $f_j$ with $X$ in its scope, i.e.\ having $p(X{=}x_i|f_j)=0$ for any factor, remove $x_i$ from the domain $\text{dom}(X)$ in all factors, and remove all probability table entries from the system that allows for $X{=}x_i$.
	
	
	\vspace{0.5em}$\hspace{-1em}$\textbf{Propagating local redundancies:}\, 
	For any two factors, $f_i$ and $f_j$, which have common variables, say $\{A, B, \ldots \}$, any zero outcome in $f_i$, i.e.\ $p(A{=}a, B{=}b, \ldots|f_i)=0$, should also be zero for $f_j$, i.e.\  as $p(A{=}a, B{=}b, \ldots|f_j)=0$. The PGM inference from Section~\ref{sec:basig_pgm} is, in fact, an algorithm to enforce this relationship, as Dechter~\cite{dechter2010on} proved this to be an algorithm for generalised arc consistency.

	We can now combine these techniques along with our merging techniques to build a PGM-based CSP solver.
	
	\subsection{The purge-and-merge procedure}
	Having outlined all the building blocks needed for purge-and-merge, we can now describe the overall concept in more detail.
	
	We start our model with factors of small-variable scopes by using the CSP clauses directly. We then incrementally transition towards a model with larger-factor scopes by clustering and merging factors. More specifically, we start with a PGM of low-factor scopes, purge redundancies from this model, progress to a model of larger-factor scopes, and purge some more redundancies. We continue this process until our PGM is tree structured and thus yields an exact solution to the CSP.
	
	
	This incremental-factor growth procedure dampens the exponential blow-up of the probability tables and allows the model to incrementally reduce the problem space. The full procedure is outlined in Algorithm~\ref{alg:purgeandmerge}.
	
	\begin{algorithm}[h!]
		\caption{\ Purge-and-Merge}
			{\bf Input:} set of factors $\mathcal{F} =\{f_1, \ldots, f_n\}$. \\
			{\bf Output:} solved variables $\mathcal{X}_s = \{x_i, \ldots \}$ and
			solved factors 
			\\${}$\hspace{3.49em} $\mathcal{F}' = \{f'_i, \ldots \}$.
		\vspace{-0.5em}
		\\
		\rule{\columnwidth}{0.45pt}
		\begin{algorithmic}[1]
			\State $\mathcal{X}_s = \{\}$
			\While{return conditions not met}
			\State $\hat{H}_\tau$ := an increasingly larger threshold
			\State {\color{blue1}// Factor clustering from Algorithm~\ref{alg:clustering}: }
			\State $\mathds{C}$ := Factor-Clustering($\mathcal{F}$, $\hat{H}_\tau$)
			\State $\mathcal{F}'$ := $\{(\prod_{f_i \in \mathcal{C}} f_i)\,$ for each $\mathcal{C} \in \mathds{C} \}$
			\State {\color{blue1}// LTRIP and LBU from Section~\ref{sec:basig_pgm}:}
			\State $\mathcal{G}$ := LTRIP($\mathcal{F}'$)
			\State $\mathcal{F}'$ := Loopy-Belief-Update($\mathcal{G}$)
			\State {\color{blue1}// Domain reduction from Section~\ref{sec:factorpurging}:}
			\State Reduce-Domains($\mathcal{F}'$)
			\State {\color{blue1}// Variable reduction from Section~\ref{sec:factorpurging}:}
			\State $\mathcal{X}$ := Reduce-Variables($\mathcal{F}'$)
			\State $\mathcal{X}_s$ := $\mathcal{X}_s \cup \mathcal{X}$\ \, {\color{blue1}// add solved variables}
			\If {$\mathcal{G}$ is a tree structure}
			\State return $\mathcal{X}_s$, $\mathcal{F}'$
			\EndIf
			\State $\mathcal{F}$ := $\mathcal{F}'$
			\EndWhile
		\end{algorithmic}\label{alg:purgeandmerge}
	\end{algorithm}

	\subsection{Algorithmic consistency}
	As a final reflection on purge-and-merge, we state that all steps taken by this algorithm are correct and result in a consistent CSP solver. Constraint satisfaction falls in the problem class of NP-complete~\cite{constraintnetworks}, and any general CSP solvers such as purge-and-merge must therefore be at least NP-complete.
	
	Dechter~\cite{dechter2010on} proves that loopy-belief propagation performed on cluster graphs with flattened tables and hard constraints reduces to generalised arc consistency. They also prove that zero-belief conclusions converge within $O(n \cdot t)$ iterations of loopy-belief propagation and that the algorithm results in a complexity of $O(r^2 t^2 \log t)$ -- where $n$ is the number of nodes in the cluster graph and $t$ bounds the size of each sparse table. These metrics are, however, not particularly useful for purge-and-merge, since the values $n$ and $t$ are expected to change as the algorithm progresses.
	
	The merging of factors is exponential in time complexity, with some merge orders performing better than others~\cite{koller}. Finding an optimal merge order is unfortunately an NP-complete problem and is as difficult as the actual inference~\cite{koller}. Some algorithms such as variable elimination can aid in this process. With variable elimination, the merging order is determined according to the marginalisation of each variable from the factors and the predicted effect it will have on the system as a whole~\cite{koller}. It should be noted that factor multiplication is commutative and any merge order converges to the same solution.
	
	Since purge-and-merge is a combination of belief propagation and factor merging, the full algorithm is bounded by factor merging and is thus also NP-complete. Purge-and-merge does, however, mitigate between loopy belief propagation and factor merging with the aim of preventing exponential blow-up in the factor-merging process.

	In the next section we investigate the performance of this procedure by applying it to a large number of CSP puzzles.	
	
	\section{Experimental study of purge-and-merge}\label{sec:experiments}
	In this section we investigate the reliability of the purge-and-merge technique by solving a large number of constraint-satisfaction puzzles. To compare results, we include Sudoku datasets used in other constraint-satisfaction PGM reports~\cite{MoonT, GoldbergerJ, KhanS, BaukeH, LakshmiA, streicher}, as well as the most difficult Sudoku puzzles currently available~\cite{champagne}.
	
	\subsection{Puzzle data sets}
	The Sudoku community has developed a large database of the hardest 9x9 puzzles~\cite{champagne} known to literature. The result is a unique collaborative research effort, spanning over a decade, using the widely accepted criterion of the Sudoku explainer rating (SER). This rating is applied by solving a puzzle using an ordered set of 96 rules ranked according to complexity. From the combination of rules required to solve the puzzle, the most difficult one of these rules is used as a hardness measure. The validity of SER as a difficulty rating is discussed by Berthier~\cite{csppuzzle}. They found that the SER rating highly correlates to external pure-logic ratings and can thus be used as a proxy for puzzle complexity. They do note that SER is not invariant to puzzle isomorphisms, i.e. two puzzles from the same validity-preserving group~\cite{russell2006mathematics} can result in two different ratings.
	
	In addition to the above set, we also compiled a database of constraint puzzles from various other sources to be used as tests. We verified each puzzle to have valid constraints using either PicoSat~\cite{picosat} or Google's OR-Tools~\cite{ortools}. All puzzles are available on GitHub~\cite{streichergithub}, and were sourced as follows:
	\begin{itemize}
		\item
		1000 \textbf{Sudoku} samples from the SER-rated set of the most difficult Sudoku puzzles, curated by Champagne~\cite{champagne},
		
		\item
		all 95 \textbf{Sudokus} from the Sterten set~\cite{sterten} used in Strei\-cher~\cite{streicher} and in Khan~\cite{KhanS},
		
		\item
		all 49151 \textbf{Sudokus} with 17-entries from the Royle's 2010 set~\cite{RoyalG} (note that an older subset of roughly 350000 puzzles was available to Goldberg~\cite{GoldbergerJ} and Bauke~\cite{BaukeH}),
		
		\item
		10000 \textbf{Killer Sudokus} from \url{www.krazydad.com} (labelled according to five difficulty levels).
		
		\item
		4597 \textbf{Calcudokus} of size $9\mkern-3mu \times \mkern-3mu 9$ from \url{www.menneske.no},
		
		\item
		6360 \textbf{Kakuro} puzzles from \url{www.grandgames.net},
		
		\item
		2340 \textbf{Fill-a-Pix} puzzles from \url{www.grandgames.net}, and
		
		\item
		a mixed set of fairly high difficulty, containing one of each of the above puzzle types.
	\end{itemize}

	\subsection{Clustering metrics}
	Section~\ref{sec:factorclustering} listed three metrics for the purge-and-merge procedure, namely (1) variable overlap, (2) upper-bound shared entropy, and (3) the gravity method. In order to select a well-adapted clustering method, we compared these three metrics on the Champagne data set. Our approach was to allow purge-and-merge to run for $10s$ under the different clustering conditions, and to then report on the largest table size for that run.
	
	Under the naive variable overlap metric, none of the puzzles came to convergence. When investigating further and re-running the first 10 puzzles without time restriction, they all ran out of physical memory. This is not surprising, as this metric does not account for the domain sizes of the variables, which can have a considerable impact on table size.
	
	Of the remaining metrics, the gravity method had a $100\%$ convergence rating within the $10s$ threshold, whereas upper-bound shared entropy had a $53\%$ rating. Compared to upper-bound shared entropy, the gravity method also resulted in a smaller maximum table size in $74.7\%$ of cases. A histogram representing the maximum table size for each run can be seen in Figure~\ref{fig-histogram}.
	
	\begin{figure}[h!]
		\centering
		\includegraphics[width=1\columnwidth]{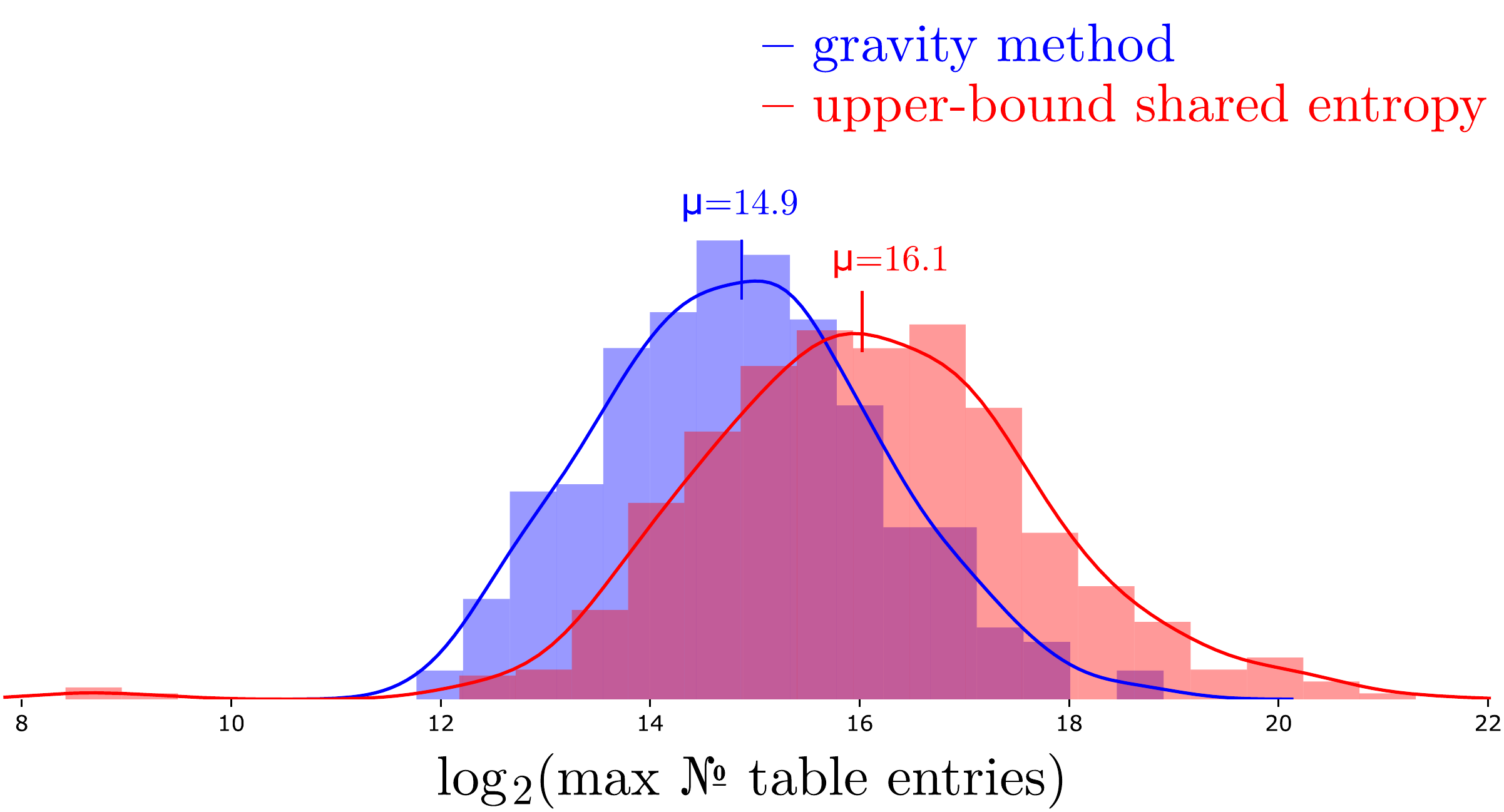}
		\caption{The maximum table produced in a purge-and-merge run using two different clustering methods: the upper-bound shared entropy method and the gravity method. All tests are run on the Champagne data set. Only runs where both methods resulted in a convergence within $10s$ are displayed.}\label{fig-histogram}
	\end{figure}
	
	Since the gravity method performed better than the other metrics, we opted to use it in all further purge-and-merge processes.
	
	\subsection{Purge-and-merge}
	All tests were executed single-threaded on an Intel$^\text{\textregistered}$  Core$^\text{TM}$ i7-3770K, with rating 3.50GHz and 4 cores / 8 threads in total. The purge-and-merge algorithm is available on GitHub~\cite{streichergithub} as a Linux binary with a command-line interface for running any of the puzzles used in our tests.

	\begin{figure}[h!]
		\centering
		\includegraphics[width=1\columnwidth]{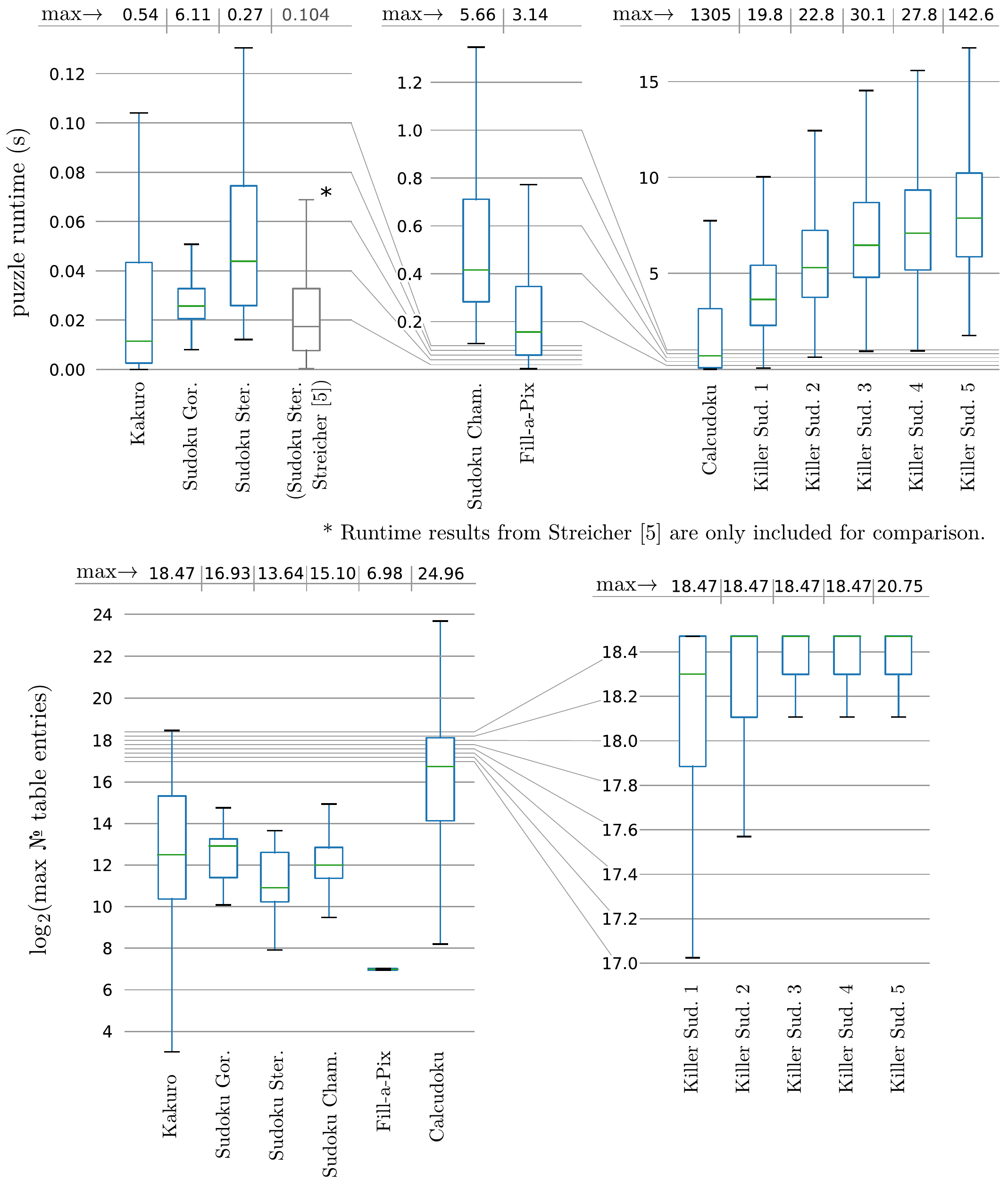}
		\caption{Runtime and size metrics for the purge-and-merge approach. The size metric indicates maximum entropy for any given factor during a purge-and-merge run, that is $\log_2(\text{maximum factor entries})$. The different Killer Sudoku sets are split according to reported difficulty, and the $1.4\%$ of unsuccessful Calcudoku runs are not included in these plots.
		}\label{fig-plotresults}
	\end{figure}
	
	Purge-and-merge can solve all the Sudoku, Killer Sudoku, Kakuro and Fill-a-Pix puzzles we have provided. In the case of Calcudoku, $1.4\%$ of the puzzle instances reached the machine's physical RAM limitation of $32$Gb. This indicates that purge-and-merge deals better with large numbers of small factors, as with Kakuro, rather than a small or medium number of large factors, as with Calcudoku and Killer Sudoku. Runtime metrics for the various puzzles can be seen in Figure~\ref{fig-plotresults}.
	
	
	Compared to the other available PGM approaches, purge-and-merge is the only method to achieve a $100\%$ success rate with all the Sudoku puzzles it encountered. Moreover, purge-and-merge was tested on more cases than what is reported in any of the comparable literature.
	
	If we compare purge-and-merge to Streicher's~\cite{streicher} results for the Sterten~\cite{sterten} set, purge-and-merge is slower (see Figure~\ref{fig-plotresults}). However, the approach by Streicher~\cite{streicher} is equivalent to a single ``purge'' step in purge-and-merge. The full purge-and-merge method obtained a success rate of $100\%$, whereas Streicher~\cite{streicher} reported a success rate of only $36.8\%$.
	
	In comparison with the other Sudoku PGM literature, Streicher~\cite{streicher} is the only other PGM approach that ensures the full CSP solution space is preserved (i.e. no valid solutions are lost). Additionally, purge-and-merge allows the scope of the solver to increase up to the point where only the solution space is left.
	
	The solution space is not preserved in the PGM approaches of Khan~\cite{KhanS}, Goldberger~\cite{GoldbergerJ} and Bauke~\cite{BaukeH}; instead, they use sum-product BP to seek out a single likely solution from the problem space. Khan~\cite{KhanS} provides us with a comparison between these three approaches, as shown in Table~\ref{tab-comparison}. From this table, it is clear that these reported PGM approaches are not well suited for Sudoku puzzles of medium and higher difficulty.
	
	\begin{table}
		\resizebox{\columnwidth}{!}{%
			\begin{tabular}{|c|c|c|}
				\cline{1-3}
				Research &  Approach & \makecell{Reported accuracy \\ for $9\times 9$ Sudokus}  \\ \cline{1-3}
				
				\multicolumn{1}{ |c  }{\multirow{2}{*}{Bauke~\cite{BaukeH}} } &
				\multicolumn{1}{ |c| } {Sum-Product} & $53.2\%$      \\ 
				\multicolumn{1}{ |c  }{} &
				\multicolumn{1}{ |c| } {Max-Product} & $70.6\%$      \\ \cline{1-3}

				\multicolumn{1}{ |c  }{\multirow{3}{*}{Goldberger~\cite{GoldbergerJ}} } &
				\multicolumn{1}{ |c| } {Sum-Product} & $71.3\%$   \\ 
				\multicolumn{1}{ |c  }{}                        &
				\multicolumn{1}{ |c| } {Max-Product} & $70.7\% - 85.6\%$   \\ 
				\multicolumn{1}{ |c  }{}                        &
				\multicolumn{1}{ |c| } {Combined Approach} &  $76.8\% - 89.5\%$  \\ \cline{1-3}

				\multicolumn{1}{ |c  }{\multirow{2}{*}{Khan~\cite{KhanS}} } &
				\multicolumn{1}{ |c| } {Khan with 40 iterations} & $70\%$     \\ 
				\multicolumn{1}{ |c  }{} &
				\multicolumn{1}{ |c| } {Khan with 200 iterations} & $95\%$      \\ \cline{1-3}
				
				\multicolumn{1}{ |c  }{\multirow{1}{*}{This paper}} &
				\multicolumn{1}{ |c| }{ Purge-and-merge} & $100\%$   \\ \cline{1-3}
			\end{tabular}
		}
		\caption{The success rate of various PGM approaches on Sudoku puzzles, originally compiled by Bauke~\cite{BaukeH}. Note that this applies to puzzles far easier than our expansive set, which also includes the current most difficult Sudoku set~\cite{champagne}.} \label{tab-comparison}
	\end{table}

	\section{Comparison to the ACE system}
	
	The ACE system~\cite{acewebsite} is a related system for solving constraint-satisfaction PGMs. ACE works by compiling Bayesian networks and other factor-graph representations into an arithmetic circuit, which can then be used to answer queries about the input variables. 
	
	ACE focuses on the marginals of the variables of the system and not on finding the joint distribution of the system. This distinction is important -- purge-and-merge produces all the solutions to a CSP, whereas ACE will only report on the marginal of each variable. 
	
	To illustrate, if we take the first 10 puzzles from the Champagne data set and arbitrarily remove a known entry, purge-and-merge finds $426, 380, 917, 799, 77, 476, 454, 1754, 777$ and $796$ answers for each puzzle respectively. ACE, on the other hand, only reports on the domain of each unknown variable and is therefore not able to find any valid solution.
	
	ACE approaches the problem in two stages. It first compiles a network along with its unknown variables into an arithmetic circuit. Then it uses the compiled network to answer multiple queries with respect to the unknown variables. Note that a single ACE circuit to represent all Sudoku puzzles is too large to fit in 32GB of memory, due to the large number of possible solutions $\approx 6.67 \times 10^{21}$~\cite{numberofsudokus}.
	
	To compare ACE with purge-and-merge, we parsed all the Champagne puzzles into ACE-compliant structure, and then compiled each puzzle into an ACE circuit. In plotting the result in Figure~\ref{fig-histogram_ace}, we discarded the loading and query times, since all evidence was already incorporated into the structure.
	
	\begin{figure}[h!]
		\centering
		\includegraphics[width=1\columnwidth]{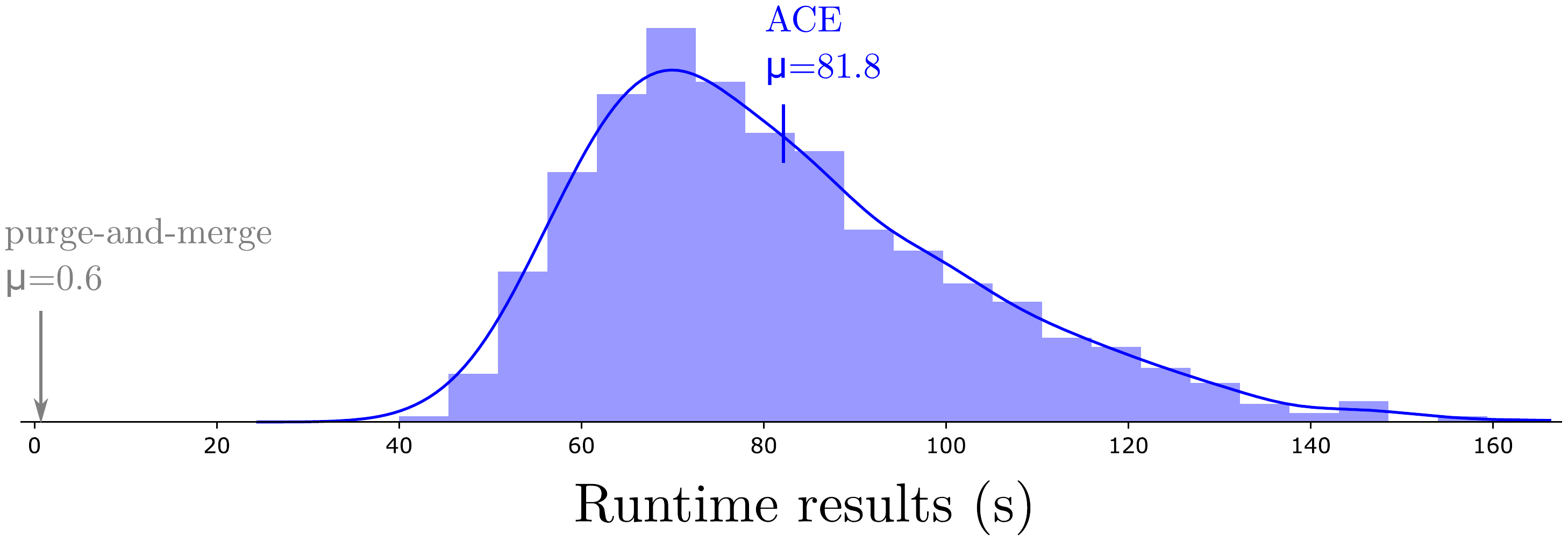}
		\caption{ACE runtime on the Champagne dataset. Only the network's compile times were recorded since the query times were negligible. For comparison, the average purge-and-merge runtime is indicated.}\label{fig-histogram_ace}
	\end{figure}

	\section{Conclusion and future work}
	
	In general, the factors in a PGM can be linked up in different ways, resulting in different graph topologies. If such a graph is tree-structured, inference will be exact. However, more often than not the graph structure will be loopy, which results in inexact inference~\cite{koller}. Transforming a loopy graph into a tree structure, unfortunately, is not always feasible -- in all but the simplest cases the resultant hyper-nodes will exponentially blow up to impractical sizes. Hence we are usually forced to work with message passing on a loopy graph structure. 
	
	The ubiquitous factor graph is the structure most frequently encountered in the literature -- its popularity presumably stems from its simple construction. Previous work has shown that inference on factor graphs is often inferior to what can be obtained with more advanced graph structures such as cluster graphs~\cite{streicher}. Nodes in cluster graphs typically exchange information about multiple random variables, whereas a factor graph is limited to sending only messages concerned with single random variables \cite[p406]{koller}. The LTRIP algorithm~\cite{streicher} enables the automatic construction of valid cluster graphs. Despite their greater potency, however, they might still be too limited to cope with complicated relationships~\cite{streicher}. 
	
	In our current work, we extend the power of cluster graphs by dynamically reshaping the graph structure as the inference procedure progresses. Semantic constraints discovered by the inference procedure reduce the entropy of some factors. Factors with high mutual attraction can then be merged without necessarily suffering an exponential growth in factor size. The LTRIP algorithm reconfigures a new structure that becomes progressively more sparse over time. When the graph structure morphs into a tree structure, the process stops with an exact solution. We refer to this whole process as purge-and-merge. 
	
	Purge-and-merge is especially useful in tasks that, despite an initially huge state space, ultimately have a small number of solutions. By hiding zero-belief conclusions from memory, purge-and-merge can perform calculations on sub spaces within an exponentially large state space.
	
	The purge-and-merge approach is not suited to tasks where the number of valid solutions would not fit into memory, as this would preclude a sufficient reduction in factor entropy. However, as the above results show, purge-and-merge enabled us to solve a wide range of problems that were previously beyond the scope of PGM-based approaches. 
	
	In comparison with ACE, we find purge-and-merge more suited to constraint-satisfaction problems with multiple solutions, as well as puzzles with a problem space too large to be compiled into a single ACE network. 
	
	Our current approach relies on the increased sparsity of the resultant graphs to gradually nudge the system towards a tree structure. In future work, we intend to control that process more actively. This should result in further gains in efficiency, and it is our hope that it will conquer the couple of Calcudokus that still elude us.


	\bibliographystyle{unsrt}
	\bibliography{do-not-edit}

\begin{IEEEbiography}[{\includegraphics[width=1in,height=1.25in,clip,keepaspectratio]{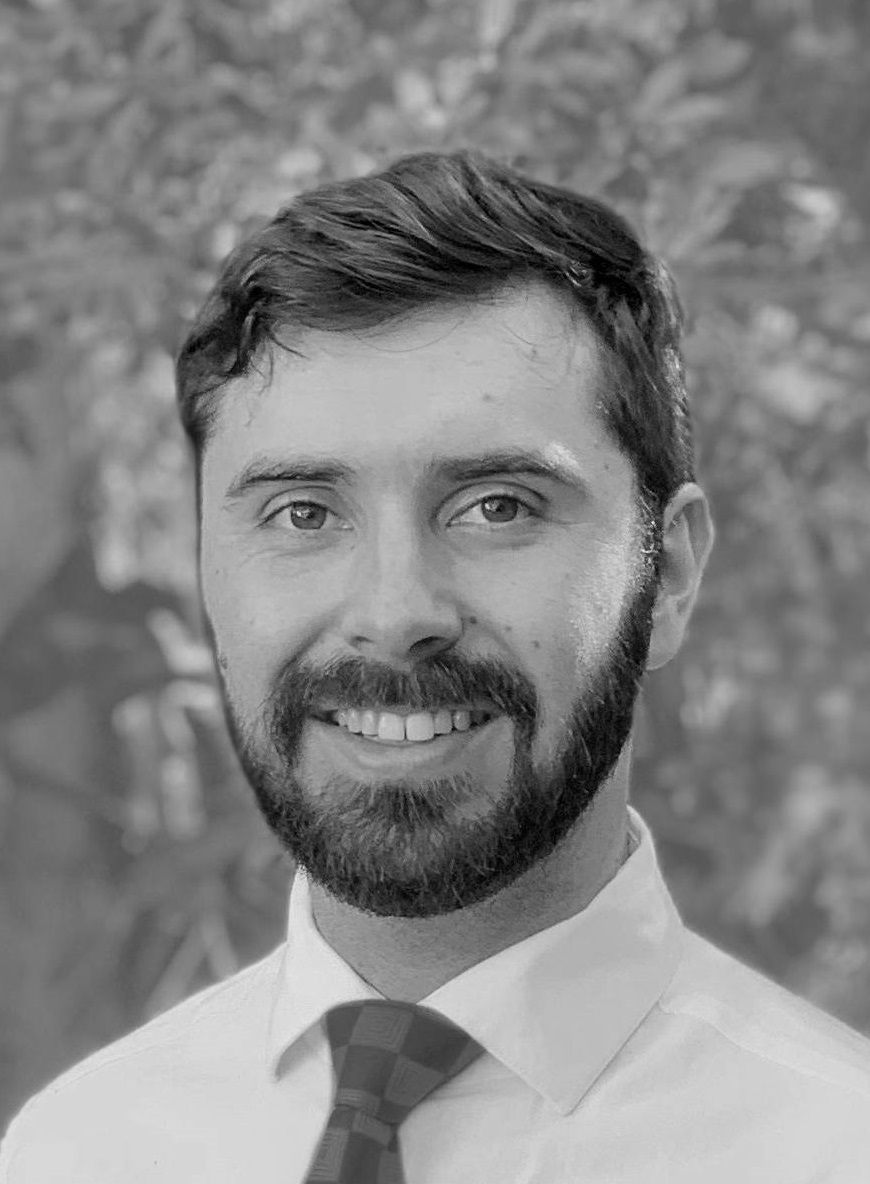}}]{Simon Streicher} 
	received his B.Eng. degree in electronic engineering in 2012 and his M.Sc. degree in applied mathematics in 2016 from Stellenbosch University, South Africa. He was a Teaching Assistant with the electronic engineering department (2016-2017) and has since enrolled for a Ph.D. in electronic engineering, while also starting a career as an actuarial consultant.
	
	His research interests include optimising probabilistic graphical model (PGM) structures and applying PGMs to structure and motion reconstruction from images, land-cover classification, and solving constraint-satisfaction puzzles.
	
	Mr. Streicher was awarded the John Todd Morrison medal for the best M.Sc. student in applied mathematics in Stellenbosch University's Faculty of Science. 
\end{IEEEbiography}

\begin{IEEEbiography}[{\includegraphics[width=1in,height=1.25in,clip,keepaspectratio]{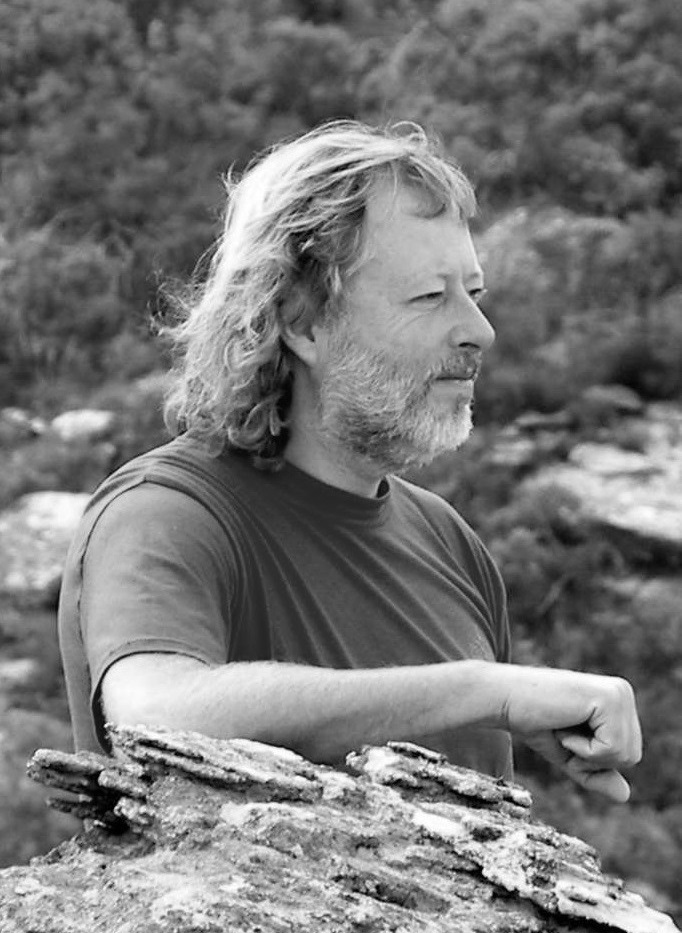}}]{Johan A. du Preez} 
	joined the Department of Electrical and Electronic Engineering at the University of Stellenbosch in 1989 after four years in the telecommunications sector. He received his Ph.D degree in electronic engineering from the University of Stellenbosch in 1998. He is active in the broader fields of signal processing and pattern recognition, with a particular research interest in developing advanced structured probabilistic models to address speech, image and text processing problems.

\end{IEEEbiography}

\EOD
\end{document}